\documentclass[11pt,a4paper]{article}
\usepackage[hyperref]{emnlp2020}
\usepackage{times}
\usepackage{latexsym}
\usepackage{multirow}
\usepackage[mathscr]{euscript}

\usepackage{graphicx}
\usepackage{booktabs,subcaption,amsfonts,dcolumn}

\usepackage{microtype}  
\usepackage{amsmath}

\aclfinalcopy 

\title{Can RNNs trained on harder subject-verb agreement instances still perform well on easier ones?}

\author{
Hritik Bansal\thanks{$^*$Equal Contribution} \textsuperscript{ 1}, Gantavya Bhatt\textsuperscript{$*$1, 2},
and Sumeet Agarwal\textsuperscript{1}\\
  \textsuperscript{1} Indian Institute of Technology Delhi\\
  \textsuperscript{2} University of Washington Seattle\\
  \texttt{hbansal10n@gmail.com} \\
  \texttt{gbhatt2@u.washington.edu}\\
  \texttt{sumeet@iitd.ac.in} 
}
\date{}

\begin{document}
\maketitle
\begin{abstract}

Subject-Verb Agreement has been a focus of investigation into the syntactical abilities of RNN language models. Previous work suggests that RNNs trained on natural language corpora can capture number agreement well for simple sentences but perform less well when sentences contain {\em agreement attractors}: intervening nouns between the verb and the main subject with grammatical number opposite to the latter. This suggests these models may not learn the actual syntax of agreement, but rather infer shallower heuristics such as `agree with the recent noun'. In this work, we investigate RNN models with varying inductive biases trained on selectively chosen `hard' agreement instances, {\em i.e.}, sentences with \emph{at least one agreement attractor}. For these the verb number cannot be predicted using a simple linear heuristic, and hence they might help provide the model additional cues for hierarchical syntax. If RNNs can learn the underlying agreement rules when trained on such hard instances, then they should generalise well to other sentences, including simpler ones. However, we observe that several RNN types, including the ONLSTM which has a soft structural inductive bias, surprisingly fail to perform well on sentences without attractors when trained solely on sentences with attractors. We analyse how these selectively trained RNNs compare to the baseline (training on a natural distribution of agreement attractors) along the dimensions of number agreement accuracy, representational similarity, and performance across different syntactic constructions. Our findings suggest that RNNs trained on our hard agreement instances still do not capture the underlying syntax of agreement, but rather tend to overfit the training distribution in a way which leads them to perform poorly on `easy' out-of-distribution instances. Thus, while RNNs are powerful models which can pick up non-trivial dependency patterns, inducing them to do so at the level of syntax rather than surface remains a challenge.



\end{abstract}

\section{Introduction}
\label{sec: Introduction}

\textbf{Subject-verb agreement (SVA)} is a phenomenon where the \emph{main subject} agrees in grammatical number with its \emph{associated verb}, oblivious to the presence of any other noun phrase in the sentence. An example is:

\begin{enumerate}
    \item \label{incorr} *The \textbf{keys} to the \underline{cabinet} \textbf{is} on the table.
    \item \label{corr} The \textbf{keys} to the \underline{cabinet} \textbf{are} on the table.
\end{enumerate}

The main noun and the associated verb are in bold. Intervening nouns are underlined, and * denotes a grammatically incorrect sentence. In the above example, the number of the main verb \emph{are} (plural) has to agree with the number of the main subject \emph{keys} (plural). Here, the intervening noun \emph{cabinet} has the opposite number (singular) to that of the main subject. Such intervening nouns are referred to as \textbf{agreement attractors} \cite{bock1991broken}. In natural language sentences, there can be any number of intervening nouns behaving as either agreement attractors or non-attractors (nouns with the same number as the main noun).

Previous work \cite{linzen2016assessing, marvin2018targeted, mccoy2018revisiting, kuncoro2019scalable, noji-takamura-2020-analysis, hao2020attribution} assessing the ability of RNN Language Models (LMs) to capture syntax-sensitive dependencies via SVA tasks has found that they often do quite well, despite lacking explicit tree structure. However, it is still not clear if good performance on such tasks is necessarily a result of the RNN's ability to capture the underlying syntax, and this is the question we seek to further investigate here.

\begin{figure}[h!]
  \centering
    \includegraphics[scale=0.5]{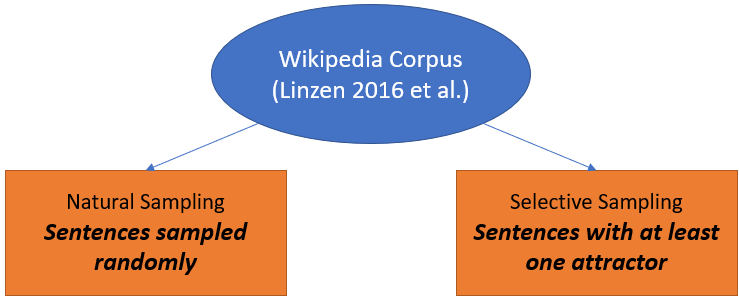}
  \caption{Dataset description. As structure-insensitive RNN models suffer from agreement attraction errors, our \textit{selectively sampled} dataset is syntactically challenging for such sequential models, for whom surface-level heuristics such as `agree with recent noun' are efficiently available. Hence, training RNNs on such datasets might induce them to capture hierarchical relations rather than learning shallower heuristics.}
  \label{fig:dataset}
\end{figure}

For a given learning task, there may be multiple hypotheses predictive of the training set labels; the learning model's \emph{inductive bias} can play a key role in selecting one hypothesis over another.
To account for the varying inductive biases that different RNN models might encode, we look at multiple architectures -- LSTM, GRU, ONLSTM, and Decay RNN (\S \ref{sec: Architectures}). \citet{mccoy2020does,mccoy2018revisiting} showed that hierarchical bias in the models, as well as the inputs, helps to generalize to unseen sentences. On the other hand, \citet{chaves-2020-dont} and \citet{sennhauser2018evaluating} provide evidence that LSTM models are more likely to learn surface-level heuristics, such as agreeing with the most recent noun, than the underlying grammar. Following \citet{mccoy2018revisiting} who show that training on syntactically rich sentences with agreement information increases the probability of good syntactic generalization, we experiment with training RNN models on sentences with at least one attractor -- \emph{Selective Sampling} (Figure \ref{fig:dataset}). Thus, the sentences in the selectively chosen dataset are syntactically richer than naturally sampled sentences, and can hopefully impart additional hierarchical cues which prevent models from relying on simple linear heuristics.

We test the hypothesis that if these models were to capture the correct grammatical structure from syntactically rich input, then they would generalize well out-of-distribution (OOD), {\em i.e.}, when tested on sentences without attractors having been trained solely on sentences with at least one attractor. Since human learners are known to have frequent access to simple structures, our learning setup itself is not realistic for human language acquisition; but the aim is to see whether it can push RNN models towards more human-like syntactic generalisation.

In addition to the examining the accuracies of these models on natural language sentences, we perform a Representation Similarity Analysis (RSA) in \S \ref{Analysis of representations}, and find that their inductive biases appear to be overridden by the training distribution, extending the findings of \citet{mccoy2020does}. 

To evaluate the effect of the choice of training data on the ability to capture SVA on a variety of syntactic constructions, we perform Targeted Syntactic Evaluation (TSE) as proposed by \citet{marvin2018targeted}, along with surprisal analysis (\S \ref{Targeted Syntactic Evaluation},\ref{Surprisal}). We find that training on selectively sampled `hard' instances improves the model's performance on difficult constructions (sentences with clauses between the main noun and the main verb), but degrades their performance on simpler constructions. Our analysis indicates that even when RNNs are able to capture non-trivial dependency patterns, they fail to perform well on easier dependency relations (instances without agreement attractors). This suggests that RNNs tend to efficiently learn surface-level heuristics from the training distribution rather than gleaning the actual syntactic rules. 

Our major outcomes are the following:
\begin{itemize}
    \item  We show that despite providing strong hierarchical cues via a selectively sampled training set (Figure \ref{fig:dataset}), RNNs, including the ONLSTM (which has a soft hierarchical inductive bias), do not generalize to an unseen configuration of intervening nouns.
    \item We observe that training data effectively override the model's structural bias, as our model representations clearly cluster by the type of training distribution. (Figure \ref{fig:RSA}).
    \item We verify that our findings are consistent across multiple learning paradigms, self-supervised language modeling and supervised grammaticality judgment, as well as varied test sets, natural and constructed (Tables \ref{tab:nat}, \ref{att}).
\end{itemize}

\section{Related Work}
\label{sec: Related Work}

Prior work by \citet{linzen2016assessing, gulordava2018colorless, marvin2018targeted, tran-etal-2018-importance} have shown that language models RNNs can capture hierarchical information from natural language to solve SVA tasks. However, it is still not clear if the models are necessarily capturing underlying syntactic rules to make a good prediction. 

\citet{mccoy2018revisiting} conclude that GRU with attention generalizes hierarchically despite the absence of hierarchical inductive bias on question formation task. Additionally, they find that training on agreement language helps in inducing syntactic bias. However, they could not conclude what these RNNs capture which gives rise to non-trivial performance. A subsequent study by \citet{mccoy2020does} shows that hierarchical generalization is only plausible by providing explicit hierarchical inputs (annotated parse information) to the model having an explicit hierarchical inductive bias (tree-based models). Similar observations have been made by \citet{kuncoro-etal-2018-lstms} and \citet{wilcox2020structural}, suggesting that structural supervision improves syntactic generalization in neural language models. In this work, we analyze if exposure to `hard' agreement instances, which cannot be modelled by inferring readily-available heuristics, can help RNNs to improve their syntactic abilities.
 

\citet{van-schijndel-etal-2019-quantity} showed that neural LMs lag far behind humans on Targeted Syntactic Evaluation (TSE), even when trained with a large corpus and increased model capacity. However, by virtue of high variance in the performance of LMs on TSE across different random seeds and hyperparameter tuning, \citet{kuncoro2019scalable} achieved better accuracy in the same setting. \citet{noji-takamura-2020-analysis} and \citet{kuncoro2019scalable} came up with methodologies to improve performance on TSE, via contrastive learning and knowledge distillation from models with explicit grammar induction \cite{dyer2016recurrent} respectively. It has been observed that these models have trouble capturing agreement in sentences with center embeddings, especially agreement across object relative clauses (RCs) \cite{noji-takamura-2020-analysis, mueller-etal-2020-cross}. In our work, we show that syntactically rich training sentences can also improve performance on such instances substantially (Table \ref{TSE table}, Appendix Table 5).

Studies on linguistic phenomena such as filler-gap dependencies  \cite{chaves-2020-dont} and assessing the ability of LSTMs to capture the rules of context-free grammars  \cite{sennhauser2018evaluating} argue that LSTMs learn shallower heuristics from the dataset rather than acquiring the underlying linguistic rules. In this work, we show that even a model with implicit hierarchical inductive bias (the ONLSTM) is behaviourally close to a vanilla LSTM in terms of the learned representations, for both supervised and self-supervised learning settings. 

\section{Architectures}
\label{sec: Architectures}
In this work, we conduct our experiments on four recurrent schemes -- LSTM \cite{hochreiter1997long}, GRU \cite{cho-etal-2014-learning}, Decay RNN (DRNN) \cite{bhatt-etal-2020-much}, and ONLSTM \cite{shen2018ordered}. The governing equations of these architectures are mentioned in Appendix \S A.1. The ONLSTM, unlike the other models under consideration, is a recurrent network with soft hierarchical inductive bias. The DRNN is a recurrent network without any gating mechanism that imposes biologically-inspired constraints on the neurons. Amongst all these RNN types, the DRNN is the least complex, and has been shown to outperform vanilla RNNs and perform on par with other gated networks on grammaticality judgment tasks \cite{bhatt-etal-2020-much}. 

\begin{table}[ht]
\centering
\resizebox{\linewidth}{!}{%
\begin{tabular}{l | r | r}
\textbf{Property}                                    & Natural  & Selective     \\
\hline
Training sentences~                             & 97842  & 97842    \\
Ratio of Singular to Plural main nouns      & 67\%    & 45\%     \\
Ratio of Singular to Plural nouns (total)   & 79\%    & 71\%     \\
Fraction of 0 attractors                    & 93\%    & -        \\
Fraction of 1 attractors                    & 5.6\%   & 79\%     \\
Fraction of 2 attractors                    & 1.1\%   & 15\%     \\
Fraction of 3 attractors                    & 0.3\%   & 3.7\%   \\
\hline
\hline 
Testing Sentences & 157k & 157k
\end{tabular}}
\caption{Data statistics.}
\label{appen table: data analysis}
\end{table}

\section{Dataset}
\label{sec: Dataset}

We use sentences from the Wikipedia corpus made available by \citet{linzen2016assessing}. For training, we further choose two subsets from the main dataset, based on the number of attractors in each sentence (Figure \ref{fig:dataset}). The sentences without any attractor are grammatically simple and allow for out-of-distribution testing as they are not seen while training on the selectively sampled dataset. We train our models for two objectives: language modeling and binary classification for grammaticality judgment (\S \ref{Experiments}). For uniform comparison, we keep the testing set identical across the subsets of the training data. The testing set contains $157$k sentences for both the binary classifier and the LM. Table \ref{appen table: data analysis} contains a quantitative description of the datasets. 

To perform well on the selectively sampled dataset, models cannot resort to learning simple linear heuristics such as associating the main verb with the preceding noun rather than the main noun. However, such heuristics are encouraged by a naturally sampled dataset due to a heavy skew towards sentences without attractors (Table \ref{appen table: data analysis}). The tendency to capture these shallow patterns would allow models to correctly predict the verb number in most instances, but for the wrong reasons.

For the binary classifier, we augment each sentence with its corresponding counterfactual example. Augmenting with counterfactual examples has been shown to be effective in reducing the tendency of the models to capture spurious correlations (for sentiment analysis) by \citet{Kaushik2020Learning}. Consider the example from \S \ref{sec: Introduction}: 
\begin{enumerate}
    \item \label{corr} The \textbf{keys} to the \underline{cabinet} \textbf{are} on the table.
    \item \label{incorr} *The \textbf{keys} to the \underline{cabinet} \textbf{is} on the table. 

\end{enumerate}
 Sentence \ref{incorr} is the counterfactual example for the sentence \ref{corr}. Thus our dataset will consist of pairs of a sentence and its corresponding counterfactual sentence. 

Additionally, we test our models on an artificially constructed corpus with a different distribution of sentence types than the training set. As proposed by \citet{marvin2018targeted}, this helps in evaluating if the models have mastered syntax across different syntactic constructions (\S \ref{Targeted Syntactic Evaluation}).

\section{Experiments}
\label{Experiments}
We focus on evaluating the models' ability to make number agreement judgments when trained for classification (supervised) and language modeling (self-supervised). For each task, we train models (with 5 different random seeds) on both training subsets from the corpus.\footnote{Code will be made available as part of the camera-ready version.} Training settings for our experiments are mentioned in Appendix \S A.2. 
Consider the sentences from the introduction. A classifier is expected to label sentence \ref{incorr} as ungrammatical and sentence \ref{corr} as grammatical. For grammaticality judgment via a language model (LM), we train on a standard LM objective and during inference, check if our model gives a higher probability to the grammatically correct verb form conditioned on previous tokens in the sentence.

\begin{table}[ht]
\centering
    \resizebox{\linewidth}{!}{%
\begin{tabular}{l | c c  c c| c c c c} 
Training set
       & \multicolumn{4}{c|}{Natural Sampling}         & \multicolumn{4}{c}{Selective Sampling}                           \\ 
\hline
Test attractors
      & 0          & 1          & 2          & 3          & 0          & 1          & 2          & 3           \\ 
\hline
      & \multicolumn{8}{c}{LANGUAGE MODEL}                                                                    \\ 
\hline
LSTM   & \textbf{0.98} & 0.91 & 0.84 & 0.78 & 0.89 & \textbf{0.98} & \textbf{0.98} & 0.95  \\ 
ONLSTM &\textbf{ 0.98} & 0.92 & 0.86& 0.82& 0.90 & \textbf{0.98} & \textbf{0.98} & 0.95  \\ 
GRU    & \textbf{0.97} & 0.88& 0.78 & 0.73& 0.87 & \textbf{0.98} & 0.97 & 0.94 \\ 
DRNN   & \textbf{0.96} & 0.69 & 0.47 & 0.36 & 0.83 & \textbf{0.97} & 0.94 & 0.91 \\ 
\hline
      & \multicolumn{8}{c}{BINARY CLASSIFIER}                                                                 \\ 
\hline
LSTM   &\textbf{ 0.97} & 0.93 & 0.87& 0.82& 0.60& \textbf{0.98} & 0.96 & 0.97  \\ 
ONLSTM &\textbf{ 0.97} & 0.91 & 0.84 & 0.81 & 0.64 & \textbf{0.98} & 0.97 & \textbf{0.98}  \\ 
GRU    &\textbf{ 0.97} & 0.88 & 0.76 & 0.69 & 0.62 & 0.95 & 0.94& \textbf{0.96} \\ 
DRNN   & \textbf{0.97 }& 0.90& 0.81 & 0.77& 0.70 &\textbf{ 0.97 }& 0.96 & 0.96  \\
\hline
\end{tabular}}
\caption{Accuracy of RNN architectures trained as LMs and classifiers, for test instances with an increasing number of attractors between main subject and verb; variances mentioned in Appendix Table 3. The maximum accuracy for each model and training setup across attractor counts is in bold. Note that the models trained on the selectively sampled dataset are not able to generalize well OOD (sentences without attractors).}
\label{tab:nat}
\end{table}

\subsection{Performance on Natural Sentences}
\label{Performance on Natural Sentences}

Table \ref{tab:nat} shows the main results for the described experiments. For the models trained on a naturally sampled dataset, the performance degrades quite quickly with an increasing number of attractors between the subject and the corresponding verb, for both the LM and the classifier versions. However, the reduction in the accuracy with increasing attractor count for the models trained on the selectively sampled dataset is much less than with the natural sampling training. 

For the selectively sampled dataset, the sentences without attractors serve as OOD sentences, and the performance boost on \emph{in-distribution} complex sentences comes at the cost of a reduction in the accuracy of the OOD yet relatively simple sentences.
The error rate for the ONLSTM, a model with inherent tree bias, also increases when tested on the OOD sentences, and when trained for a classification objective it performs worse than the architecturally simpler Decay RNN.

This fall-off on grammatically simpler OOD samples seems counter-intuitive. We note that the increase in error rates is much greater when training the models as classifiers rather than LMs. This shows that models with supervised training for grammaticality on syntactically rich and counterfactually augmented data are still unable to capture the actual syntactic rules and appear to be learning shallower heuristics, but ones that capture more nuanced patterns than simply going by linear distance. We can infer this because while our selectively sampled subset contains sentences with at least one attractor, many (over 30\%) of the intervening nouns in these sentences are non-attractors. Hence there are sentences in which a non-attractor noun (same number as the main subject) immediately precedes the verb rather than an attractor noun. Therefore, the agreement performance (on sentences with attractors) of the models trained on this dataset cannot arise from an overly simple heuristic like disagreeing with the most recent noun, and the observed decline in OOD performance implies that less trivial heuristics are being learned which nevertheless fail to capture the actual syntax.

\subsection{Analysis of representations}
\label{Analysis of representations}

To analyze the differences in the learned internal representations among the models trained on the two subsets of the data, we perform a representation similarity analysis (RSA) \cite{laakso2000content}. We take 2000 sentences selected randomly from the test set. As we had trained each model on five different random seeds, we compare 40 models (20 for each subset) across different learning objectives. 

\begin{figure}[h!]
  \centering
  \centering
  \begin{subfigure}[b]{\linewidth}
    \includegraphics[width=\textwidth]{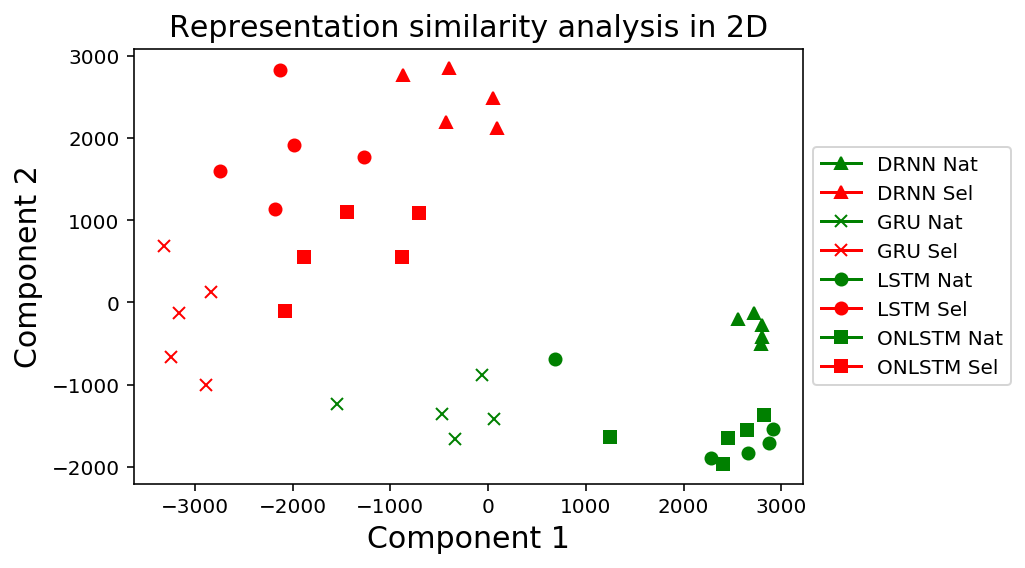}
    \caption{Binary classifier}
    \label{BC RSA}
  \end{subfigure} \hfill
  \begin{subfigure}[b]{\linewidth}
    \includegraphics[width=\textwidth]{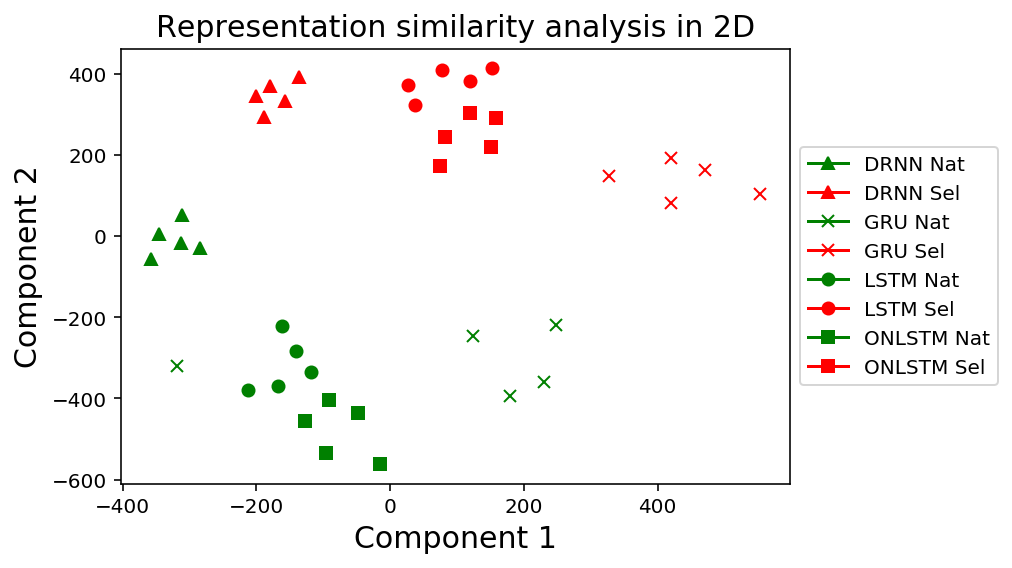}
    \caption{Language Model}
    \label{LM RSA}
  \end{subfigure}
  \caption{Representation similarity analysis of the hidden units of different RNN models (5 different seeds for each model). We observe that for both the learning objectives, one can partition the 2D space using a line that separates models trained on the two subsets of the data, natural and selective sampling.}
  \label{fig:RSA}
\end{figure}

Our major observation from Figure \ref{fig:RSA} is that the representations of models trained on different subsets are easily linearly separable in this space, for both the LM and the classifier objectives. This implies that the representation clustering is not so much based on model architecture or inductive bias, but is overridden by the training data. 

Additionally, we observe that the learned representations of the ONLSTM and LSTM are not well separable, neither for the LM nor for the classifier across training configurations. This shows that despite having a soft hierarchical inductive bias, the ONLSTM appears similar to the regular LSTM in terms of patterns captured at the representational level. Moreover, the learned representations for GRU and LSTM are well separable  This differentiates the two architectures, which are often used interchangeably on the representation level, and such differences may be arising due to the squashing phenomena in GRUs pointed out by \citet{mccoy2020does}. 

For the binary classifier (Figure \ref{BC RSA}), although we observe a little variance in the accuracy on the test set across the different seeds, the variance (spread of the points) in the projected space is substantial when compared to that of the LM. Quantitative analysis of the variance is available in Appendix \S A.5. This might be due to the existence of multiple valleys in the loss landscape for the binary classification objective, and we posit that an LM objective is more reliable when comparing the ability to capture the syntax sensitive dependencies in RNNs. In the following sections, we shall further analyze the performance of models trained with LM objective in greater detail.

\begin{table*}[ht]
\centering
    \resizebox{1\textwidth}{!}{%

\begin{tabular}{l|l|c c|c c|c c|c c} 
Subject Verb Agreement                    & \#sentences & \multicolumn{2}{c|}{LSTM}                            & \multicolumn{2}{c|}{ONLSTM}                          & \multicolumn{2}{c|}{GRU}                             & \multicolumn{2}{c}{DRNN}                             \\ 
\hline
Condition &            & \multicolumn{1}{c|}{Natural} & \multicolumn{1}{c|}{Selective} & \multicolumn{1}{c|}{Natural} & \multicolumn{1}{c|}{Selective} & \multicolumn{1}{c|}{Natural} & \multicolumn{1}{c|}{Selective} & \multicolumn{1}{c|}{Natural} & \multicolumn{1}{c}{Selective}  \\ 
\hline
Simple     & 312  & \textbf{0.99} ($\pm$0.01) & 0.86 ($\pm$0.01)          & \textbf{0.98} ($\pm$0.02) & 0.86 ($\pm$0.01)          & \textbf{0.98} ($\pm$0.01) & 0.84 ($\pm$0.04)          & \textbf{0.97} ($\pm$0.02) & 0.79 ($\pm$0.05)           \\ 
Short VP                & 3432  & \textbf{0.85} ($\pm$0.02) & 0.71 ($\pm$0.06)           & \textbf{0.88} ($\pm$0.02)  & 0.73 ($\pm$0.08)           & \textbf{0.81} ($\pm$0.03) & 0.69 ($\pm$0.04)           & \textbf{0.70} ($\pm$0.05)  & 0.66 ($\pm$0.04)            \\ 
Within ORC (A)          & 9984  & \textbf{0.79} ($\pm$0.06) & 0.63 ($\pm$0.05)           & \textbf{0.78} ($\pm$0.10)   & 0.59 ($\pm$0.06)           & \textbf{0.75} ($\pm$0.02) & 0.50 ($\pm$0.02)            & \textbf{0.7} ($\pm$0.08)  & 0.46 ($\pm$0.04)            \\ 
Within ORC (IA)         & 4032  & \textbf{0.77} ($\pm$0.06) & 0.64 ($\pm$0.06)           & \textbf{0.75} ($\pm$0.08)  & 0.59 ($\pm$0.04)           & \textbf{0.73} ($\pm$0.02) & 0.50 ($\pm$0.03)            & \textbf{0.69} ($\pm$0.06) & 0.46 ($\pm$0.05)            \\ 
Within no that ORC (A)  & 9984  & \textbf{0.73} ($\pm$0.06) & 0.61 ($\pm$0.05)           & \textbf{0.72} ($\pm$0.08)  & 0.57 ($\pm$0.07)           & \textbf{0.72} ($\pm$0.03) & 0.47 ($\pm$0.04)           & \textbf{0.63} ($\pm$0.04) & 0.45 ($\pm$0.06)            \\ 
Within no that ORC (IA) & 4032  & \textbf{0.66} ($\pm$0.04) & 0.61 ($\pm$0.05)           & \textbf{0.66} ($\pm$0.06)  & 0.56 ($\pm$0.06)           & \textbf{0.62} ($\pm$0.04) & 0.47 ($\pm$0.04)           & \textbf{0.68} ($\pm$0.06) & 0.45 ($\pm$0.06)            \\ 
Long VP                 & 520   & 0.65 ($\pm$0.03)         & \textbf{0.69} ($\pm$0.07) & \textbf{0.67} ($\pm$0.04) & \textbf{0.67} ($\pm$0.06) & 0.63 ($\pm$0.04)         & \textbf{0.65} ($\pm$0.04) & 0.56 ($\pm$0.05)         & \textbf{0.65} ($\pm$0.03)  \\ 
Across Prep Phrase (A)           & 29952 & 0.86 ($\pm$0.04)          & \textbf{0.89} ($\pm$0.03)  & \textbf{0.88} ($\pm$0.03)  & \textbf{0.88} ($\pm$0.01)  & 0.81 ($\pm$0.02)          & \textbf{0.88} ($\pm$0.02)  & 0.68 ($\pm$0.04)          & \textbf{0.83} ($\pm$0.01)   \\ 
Across Prep Phrase (IA)          & 4032  & 0.87 ($\pm$0.03)          & \textbf{0.94} ($\pm$0.02)  & 0.88 ($\pm$0.02)           & \textbf{0.95} ($\pm$0.01)  & 0.86 ($\pm$0.02)          & \textbf{0.94} ($\pm$0.01)  & 0.69 ($\pm$0.06)          & \textbf{0.91} ($\pm$0.02)   \\ 
Across SRC & 9984 & 0.81 ($\pm$0.03)          & \textbf{0.89} ($\pm$0.05) & 0.81 ($\pm$0.05)          & \textbf{0.87} ($\pm$0.02) & 0.77 ($\pm$0.05)          & \textbf{0.86} ($\pm$0.05) & 0.58 ($\pm$0.04)          & \textbf{0.80} ($\pm$0.05)  \\
Across ORC (A)          & 9984  & 0.73 ($\pm$0.10)          & \textbf{0.82} ($\pm$0.07) & 0.78 ($\pm$0.07)          & \textbf{0.84} ($\pm$0.02) & 0.72 ($\pm$0.06)         & \textbf{0.79} ($\pm$0.05) & 0.63 ($\pm$0.04)         & \textbf{0.78} ($\pm$0.05)  \\ 
Across ORC (IA)         & 4032  & 0.74 ($\pm$0.09)         & \textbf{0.84} ($\pm$0.10)  & 0.81 ($\pm$0.07)          & \textbf{0.87} ($\pm$0.02) & 0.74 ($\pm$0.08)         & \textbf{0.85} ($\pm$0.05) & 0.65 ($\pm$0.07)         & \textbf{0.86} ($\pm$0.02)  \\ 
Across no that ORC (A)  & 9984  & 0.61 ($\pm$0.04)         & \textbf{0.72} ($\pm$0.08) & 0.62 ($\pm$0.05)          & \textbf{0.78} ($\pm$0.02) & 0.60 ($\pm$0.02)          & \textbf{0.68} ($\pm$0.06) & 0.64 ($\pm$0.03)         & \textbf{0.73} ($\pm$0.02)  \\ 
Across no that ORC (IA) & 4032  & 0.66 ($\pm$0.04)         & \textbf{0.77} ($\pm$0.11) & 0.66 ($\pm$0.06)          & \textbf{0.84} ($\pm$0.03) & 0.62 ($\pm$0.04)         & \textbf{0.72} ($\pm$0.07) & 0.68 ($\pm$0.06)         & \textbf{0.83} ($\pm$0.02)  \\
\hline
Average Performance                &   104296     &           0.78 ($\pm$0.03)      &  0.78 ($\pm$0.02)              &      0.79 ($\pm$0.03)           &        0.78 ($\pm$0.01)        &   0.75 ($\pm$0.01)              & 0.73 ($\pm$0.02)                &  0.66 ($\pm$0.02)                &        0.71 ($\pm$0.02)         \\
\hline
\end{tabular}} 
\caption{Accuracy of models on targeted syntactic evaluation. Quantities in bold marks the maximum accuracy for each model across the configuration. ORC: Objective Relative Clause, SRC: Subject Relative Clause, Prep Phrase: Prepositional Phrase, VP: Verb Phrase. A/IA in the parenthesis represents an animate/inanimate main subject. Models trained on selectively sampled subset perform well on the difficult sentences, but not on the simpler ones.}
\label{TSE table}
\end{table*}

\subsection{Targeted Syntactic Evaluation (TSE)}
\label{Targeted Syntactic Evaluation}

We test how training the language models on the strategically chosen inputs impacts generalisation to different syntactic constructions. Testing on such examples lets us evaluate if our models are capturing what we intend them to capture. As the subjects can be separated from their verb by long clauses, SVA requires models to capture robust hierarchical representations. The models capturing surface level regularities of the data would not be able to perform well on these constructed examples. Sentence \ref{ORC} corresponds to the agreement across an object relative clause, while the sentence \ref{SRC} corresponds to the agreement across a subject relative clause. 

\begin{enumerate}
    \item \label{ORC} The \textbf{authors} that the \underline{chef} likes \textbf{laugh}.
    \item \label{SRC} The \textbf{authors} that like the \underline{chef} \textbf{laugh}. 
\end{enumerate}
Table \ref{TSE table} mentions the results of TSE on the LM. For each model, we observe that as the difficulty of the sentences increases, models trained on the selectively sampled dataset starts surpassing those trained on the natural dataset. Difficult sentences involve number agreement across the prepositional phrase and subject/object relative clauses. However, this improvement in the performance came at a cost where their performance on simple sentences, having agreement across short verb phrases and agreement within object relative clauses, decreases. Such sentences do not have any intervening noun between the subject and the corresponding verb.

Table \ref{att} presents the performance of the LMs on the synthetic data, for sentences with 0 or 1 agreement attractors. These findings corroborate our observations on natural language sentences -- the models trained on the selectively sampled dataset performed worse on sentences without attractors which are syntactically simpler. 

As supplementary results, we present a comparison of the LSTM LM trained on the selectively sampled subset with the techniques presented by \citet{kuncoro2019scalable} in Appendix Table 5.


\begin{table}[ht]
\resizebox{\linewidth}{!}{%
\begin{tabular}{l|c|c|c|c} 
Training set & \multicolumn{2}{c|}{ Natural} & \multicolumn{2}{c}{ Selective}  \\ 
\hline
Test attractors             & 0          & 1            & 0          & 1            \\ 
\hline
LSTM         & \textbf{0.77} {($\pm$ 0.05)} & 0.66 ($\pm$ 0.04)   & 0.63 ($\pm$ 0.04) & \textbf{0.83} {($\pm$ 0.06)}   \\ 
ONLSTM       & \textbf{0.76} {($\pm$ 0.07)} & 0.70 ($\pm$ 0.06)   & 0.60 ($\pm$ 0.05) & \textbf{0.85} {($\pm$ 0.01)}   \\ 
GRU          & \textbf{0.74} {($\pm$ 0.02)} & 0.64 ($\pm$ 0.02)   & 0.51 ($\pm$ 0.02) & \textbf{0.81} {($\pm$ 0.04)}   \\ 
DRNN         & \textbf{0.67} {($\pm$ 0.04)} & 0.44 ($\pm$ 0.04)   & 0.48 ($\pm$ 0.04) & \textbf{0.79} {($\pm$ 0.03)}  \\
\bottomrule
\end{tabular}} 
\caption{Accuracy of LMs on test instances with 0 or 1 attractors from the artificial corpus. Models trained on the selectively sampled subset do not generalize well on OOD sentences without attractors.}
\label{att}
\end{table}

\subsubsection*{Fine-Grained Analysis}
\label{Fine-Grained Analysis}

\begin{figure}[h!]
  \centering
  \begin{subfigure}[b]{\linewidth}
    \includegraphics[width=\textwidth]{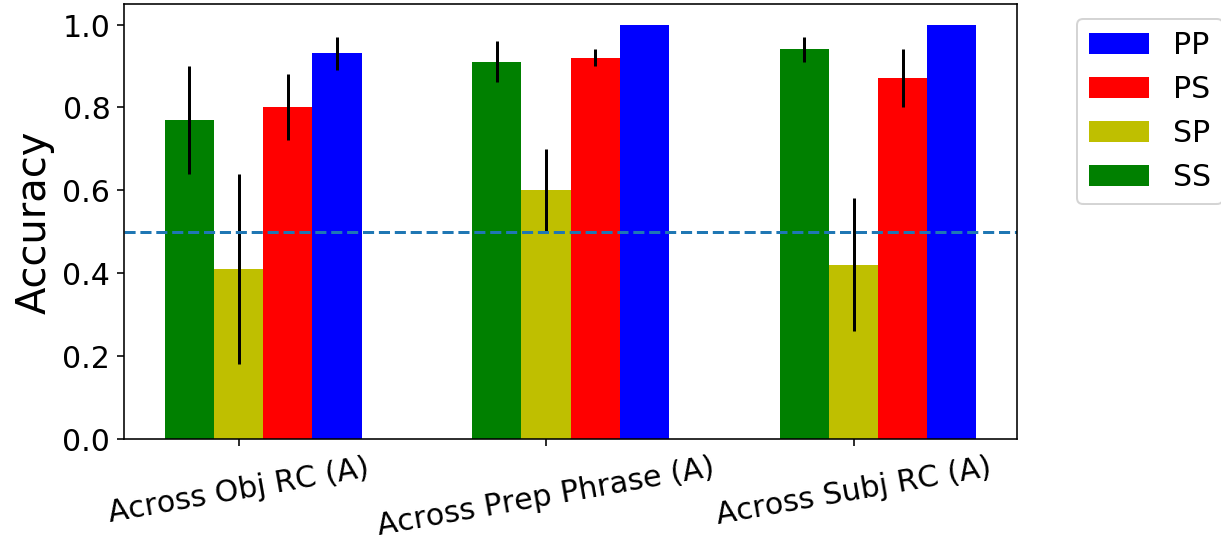}
    \caption{Accuracy: LSTM trained on naturally sampled subset}
    \label{FGA_a}
  \end{subfigure} \hfill
  \begin{subfigure}[b]{\linewidth}
    \includegraphics[width=\textwidth]{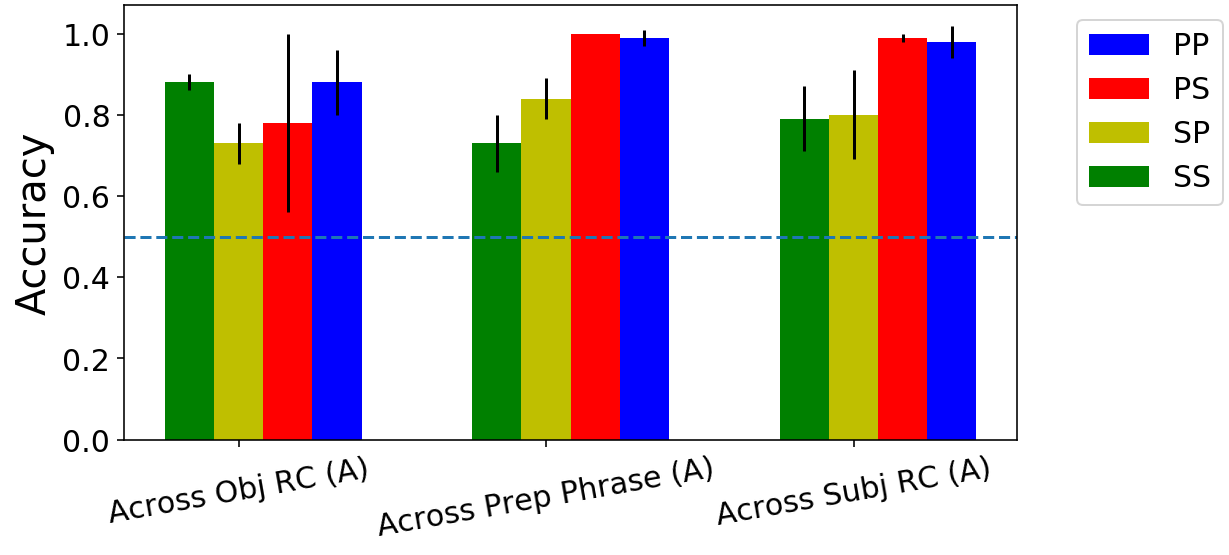}
    \caption{Accuracy: LSTM trained on selectively sampled subset}
    \label{FGA_b}
  \end{subfigure}
  \caption{Fine-grained analysis of the LSTM LM on Obj/Subj Relative Clauses and Preposition Phrases, demarcated by the inflections of the main subject and the embedded subject. P: Plural, S: Singular; thus SS denotes sentences with a singular main noun and a singular embedded subject, and likewise for the other cases.}
  \label{fig:FGA}
\end{figure}

To assess the performance of the models trained on the selectively sampled dataset, we take a closer look at constructed sentences that are structurally similar to in-distribution sentences but contain non-attractor intervening nouns rather than agreement attractors. Figure \ref{fig:FGA} depicts the performance of the LSTM LM on three agreement conditions -- across Object RC, Preposition Phrase, and Subject RC, each with an animate main noun. For sentences with two nouns, we have four possibilities corresponding to their combination of grammatical numbers. Consider the following examples for the sentences with object relative clause across the main noun and its verb ($\ast$ marks the incorrect verb). 

\begin{enumerate}
    \item \label{SS} (SS) The \textbf{author} that the \underline{minister} likes laughs/*laugh.
    \item \label{SP} (SP) The \textbf{author} that the \underline{ministers} like laughs/*laugh.
    \item \label{PS} (PS) The \textbf{authors} that the \underline{minister} likes laugh/*laughs.
    \item \label{PP} (PP) The \textbf{authors} that the \underline{ministers} like laugh/*laughs.
\end{enumerate}

Here SS denotes sentences having singular main noun and a singular embedded subject, and likewise for other cases. We observe that with our selective training, the performance on sentences with non-attractor intervening nouns (the SS/PP configurations, which are unobserved in the selectively sampled dataset) worsens for 2 out of 3 syntactic constructions -- across Preposition Phrase and Subject RC. This pattern highlights that the reduction in the performance on simple sentences may not be completely attributed to their difference in syntactic structure from training distribution.  

For all the cases under consideration presented in Figure \ref{FGA_a}, the performance of the LSTM LM is worst when the main noun is singular with the plural embedded subject (SP case). This indicates that plural attractors have a stronger attraction effect than singular attractors. Note that the naturally sampled dataset has more plural attractors while the selectively sampled one has an almost equal balance of plural and singular attractors (Table \ref{appen table: data analysis}). Thus, our findings are not due to a lack of plural attractors seen during the training phase.


For completeness, we report the fine-grained analysis accuracies for all the models under consideration in the supplementary material (Tables 6, 7, and 8).

\subsection{How confident are RNN language models?}
\label{Surprisal}

From the previous experiments, it is not clear how confident neural LMs are while making predictions about the number of the upcoming verb given the context till that verb. Surprisal at a given word is defined as $-\log p(x_n \ | \ x_{[1:n-1]})$, where $x_n$ represents the $n^{th}$ token in the sentence and $x_{[1:n-1]}$ represents the sequence of preceding tokens. We define \emph{prediction confidence} as the difference in surprisal values at the verb location between the incorrect verb and the correct verb; higher values correspond to more confident correct predictions. In this section we compare the performance of models trained on the two subsets in terms of their average prediction confidence at the verb position, evaluated on 16k simple sentences from the natural corpus (testing set) of the following form:

\begin{center}
The $<$Subject$>$ $<$Verb$>...$ 
\end{center}

Note that this amounts to evaluation on sentences involving agreement without attractors. Table \ref{surp} shows the average prediction confidence and the average ratio of predicted probabilities for the grammatical verb and the ungrammatical verb. We can see that the average probability ratio of the grammatically correct verb to the incorrect verb is much lower for the models trained on the selectively sampled subset than those trained on the naturally sampled subset. Thus, in addition to the difference between the accuracies of LMs trained on the two subsets being close to 10\% on sentences with no attractors (Table \ref{tab:nat}), models trained with selective sampling are also less confident, even when they predict correctly. Moreover, we see that the ONLSTM suffers as big a drop in prediction confidence as the regular LSTM when trained on this selectively sampled dataset, despite the former having a hierarchical inductive bias which we might have hoped would help it capture the correct syntax. 

\begin{table}[ht]
\centering
\resizebox{0.8\linewidth}{!}{%
\begin{tabular}{l|c |c|c| c} 
\multicolumn{1}{c|}{Training set} & \multicolumn{1}{c|}{Natural} & \multicolumn{1}{c|}{Selective} & \multicolumn{1}{c|}{Natural} & \multicolumn{1}{c}{Selective}  \\ 
\hline
Test metric                            & \multicolumn{2}{c|}{Prediction Confidence}    & \multicolumn{2}{c}{ \(\frac{P(grammatical)}{P(ungrammatical)}\)}  \\ 
\hline
LSTM   & \textbf{5.72} & 2.56 & \textbf{303.91} & 12.90  \\ 
ONLSTM & \textbf{5.80} & 2.64 & \textbf{331.83} & 14.08  \\ 
GRU    & \textbf{5.13} & 2.40 & \textbf{169.28} & 11.02  \\ 
DRNN   & \textbf{5.15} & 2.38 & \textbf{172.78} & 10.77  \\
\end{tabular}} 
\caption{Average prediction confidence and correct-to-incorrect probability ratio for agreement without attractors. Models trained on the selectively sampled dataset have lower confidence on non-attractor sentences than those trained on the natural dataset.}
\label{surp}
\end{table}

\subsection{Can fine-tuning help?}
\label{Can finetuning help?}

Fine-tuning has been shown to be an effective strategy to improve the syntactic robustness of RNNs. \citet{lepori-etal-2020-representations} fine-tuned their model on a small number of synthetic sentences which could not solved using word co-occurrence statistics. This eventually helped in decreasing their model's error on the number agreement task. We assess if a similar technique could improve the syntactic generalization ability of the models trained on the selectively sampled dataset. 

We fine-tune our LMs with artificially generated sentences following the rules of context-free grammar. We notice that our model's error rates did not decrease on sentences without attractor with fine-tuning. This observation may be the result of catastrophic forgetting during the fine-tuning phase \cite{jiang-etal-2020-smart, aghajanyan2020better, McCloskey1989}. Another possibility is that the RNNs have not been able to learn underlying syntactic rules from the given training distribution. Recent information-theoretic analysis by \citet{lovering2021informationtheoretic} points out that fine-tuning might not uncover new features that are not already captured during the pre-training phase. This suggests that training on `hard' instances might not have led the RNNs to capture dependency patterns at the level of syntax. See Appendix \S A.6 for fine-tuning results. 

\section{Discussion and Conclusion}
\label{Discussion}

In this work, we analyzed the effects of a strategically chosen training set with exclusively `hard' agreement instances, on neural language models and binary classifiers for grammaticality judgment. We observed that the models' inability to perform well on out of distribution (OOD) sentences, even those which would seem to be `easy' agreement instances, is consistent across variation in learning mechanism (supervised or self-supervised), innate architectural bias, and testing set -- natural or artificial sentences.

Our analysis showed that the error rates of models trained on sentences with at least one agreement attractor are higher on sentences with no attractors than on sentences with attractors, for both corpus sentences (Table \ref{tab:nat}) and artificial sentences (Table \ref{att}). This observation is counter-intuitive because the models were trained on syntactically rich sentences and yet failed to perform well on simpler sentences. Had our RNN models picked up the correct grammatical rules, we would not expect this behavior. We obtained a similar counter-intuitive result for targeted syntactic evaluation (Table \ref{TSE table}), where models trained on the selectively sampled dataset performed much better on difficult constructed sentences involving agreement across nested dependencies than on simpler sentences involving agreement within nested dependencies.

Our analysis of representations suggested that training set bias dominates over the model's architectural features or inductive bias in shaping representation learning; {\em e.g.}, there was no discernible difference between the learned representations of the ONLSTM and LSTM models. The reasons for this merit further exploration. Moreover, for the binary classifiers (Figure \ref{BC RSA}), although we observe little variance in test accuracy across different training seeds, the variance in the projected representation space is substantially greater than for LMs. Thus, we posit that an LM objective is more reliable when comparing the ability of different RNN models to capture syntax-sensitive dependencies. Additionally, multi-task learning may improve language modeling by predicting the CCG supertags of each word \cite{marvin2018targeted}, or by incorporating a contrastive loss term in the objective \cite{noji-takamura-2020-analysis}.   


We observed that the hierarchical inductive bias in the ONLSTM is not sufficient to perform well on OOD sentences. \citet{mccoy2020does} argued that an architecture with explicit tree bias, plus syntactically annotated inputs, are needed to capture syntax for sequence-to-sequence tasks. Here we show that the ONLSTM (soft tree bias) trained on a syntactically rich dataset (soft structural information) turns out to be insufficient to generalize well to OOD sentences and capture the underlying grammar. Our targeted syntactic evaluation pinpoints the cases which our models fail to capture, and improving performance on such cases is a key future direction. 

Our observations suggest that RNNs, being fundamentally statistical models, can efficiently capture the correlation of the output variable with the input as observed during training (even for relatively `hard' or non-linear linguistic dependencies) without necessarily learning the underlying hierarchical structure. This is consistent with the conclusions of \citet{sennhauser2018evaluating} and \citet{chaves-2020-dont}. Thus, we need to be cautious in inferring the ability of such models to capture syntax-sensitive dependencies. Performance on any particular kind of construction might always reflect some overfitting to it, even if it is syntactically rich or complex. Although we have here focused on training solely on hard instances as a means of hierarchical cueing, identifying the best mix of simpler and more complex sentences that a model should be exposed to for optimal generalisation performance across a variety of syntactic constructions remains a direction for future work. Broad-based testing on instances of diverse types and complexity levels is essential to the development of models that better capture the structure of human language in all its richness and variety.



\bibliographystyle{acl_natbib}
\bibliography{emnlp2020}

\clearpage
\appendix

\section{Appendix}
\subsection{Model Architectures}
\label{app: Model Architectures}
Following are the equations of the models used in this papers. `$\circ$' denotes the Hadamard product. 

\subsubsection{Long Short Term Memory (LSTM)}
Following are the equations governing the standard LSTM \cite{hochreiter1997long} with the standard notations.
\begin{equation*}
i_{t} =\sigma\left(W_{i}\left[h_{t-1}, x_{t}\right]+b_{i}\right) 
\end{equation*}
\begin{equation*}
    f_{t} =\sigma\left(W_{f}\left[h_{t-1}, x_{t}\right]+b_{f}\right)
\end{equation*}
\begin{equation*}
    g_{t} =\tanh \left(W_{g}\left[h_{t-1}, x_{t}\right]+b_{g}\right)
\end{equation*}
\begin{equation*}
    o_{t} =\sigma\left(W_{o}\left[h_{t-1}, x_{t}\right]+b_{o}\right) 
\end{equation*}
\begin{equation*}
    c_{t} =f_{t}\circ c_{t-1}+i_{t}\circ g_{t} 
\end{equation*}
\begin{equation*}
    h_{t} =o_{t}\circ \tanh \left(c_{t}\right)
\end{equation*}

\subsubsection{Gated Recurrent Unit (GRU)}
Following are the equations governing the standard GRU \cite{cho-etal-2014-learning} with the standard notations.
\begin{equation*}
r_{t}=\sigma\left(W_{r}\left[h_{t-1}, x_{t}\right]+b_{r}\right) 
\end{equation*}
\begin{equation*}
    z_{t}=\sigma\left(W_{z}\left[h_{t-1}, x_{t}\right]+b_{z}\right)
\end{equation*}
\begin{equation*}
    \tilde{h}=\tanh \left(W_{x}\left[r_{t}\circ h_{t-1}, x_{t}\right]+b_{x}\right)
\end{equation*}
\begin{equation*}
    h_{t}=z_{t}\circ h_{t-1}+\left(1-z_{t}\right)\circ \tilde{h}
\end{equation*}
\subsubsection{Ordered Neurons (ONLSTM)}
Ordered Neuron or Ordered Neuron LSTMs \cite{shen2018ordered} are recurrent schemes that have been claimed to represent hierarchical information in their representations by their $cumax$ or cumulative softmax activation. The following are the equations of Ordered Neurons with the standard notations. 

\begin{equation*}
f_{t}=\sigma\left(W_{f}\left[h_{t-1}, x_{t}\right]+b_{f}\right) 
\end{equation*}
\begin{equation*}
i_{t}=\sigma\left(W_{i}\left[h_{t-1}, x_{t}\right]+b_{i}\right) 
\end{equation*}
\begin{equation*}
o_{t}=\sigma\left(W_{o}\left[h_{t-1}, x_{t}\right]+b_{o}\right) 
\end{equation*}
\begin{equation*}
\hat{c}_{t}=\tanh \left(W_{c}\left[h_{t-1}, x_{t}\right]+b_{c}\right) \end{equation*}
\begin{equation*}
\tilde{f}_{t}=\operatorname{cumax}\left(W_{\tilde{f}}\left[h_{t-1}, x_{t}\right]+b_{\tilde{f}}\right)
\end{equation*}
\begin{equation*}
\tilde{i}_{t}=1-\operatorname{cumax}\left(W_{\tilde{i}}\left[h_{t-1}, x_{t}\right]+b_{\tilde{i}}\right)
\end{equation*}
\begin{equation*}
\omega_{t} =\tilde{f}_{t} \circ \tilde{i}_{t} 
\end{equation*}
\begin{equation*}
\hat{f}_{t} =f_{t} \circ \omega_{t}+\left(\tilde{f}_{t}-\omega_{t}\right)
\end{equation*}
\begin{equation*}
\hat{i}_{t} =i_{t} \circ \omega_{t}+\left(\tilde{i}_{t}-\omega_{t}\right)
\end{equation*}
\begin{equation*}
c_{t} =\hat{f}_{t} \circ c_{t-1}+\hat{i}_{t} \circ \hat{c}_{t}
\end{equation*}
\begin{equation*}
    h_{t} =o_{t} \circ \tanh \left(c_{t}\right)
\end{equation*}

\subsubsection{Decay RNN (DRNN)}
Decay RNN (DRNN) \cite{bhatt-etal-2020-much} is a bio-inspired recurrent baseline without any gating mechanism. Authors also show that DRNN surpasses vanilla RNNs on linguistic tasks.

\begin{equation*}
{c}^{(t)} =\left(\operatorname{Re} L U({W}){W}_{\text {dale}}\right) {h}^{(t-1)}+{U x}^{(t)}+{b}
\end{equation*}
\begin{equation*}
 {h}^{(t)} = tanh\left(\alpha {h}^{(t-1)}+(1-\alpha) {c}^{(t)}\right)
\end{equation*}

Here  $\alpha \in$ (0,1) as a learnable parameter and \(\mathbf{W}_{dale}\) is a diagonal matrix which provides biological constraints. 



\subsection{Training Settings}
\label{Training Settings}
In our experiments, we train a two-layered LM where we keep the hidden size at 650 units and the input size as 200 units. We perform standard dropout with a rate of 0.2 and the batch size 128. Optimization starts with a 0.001 learning rate for all architecture and clips the gradient if necessary.

For Binary classifiers, we use a single-layered recurrent unit, batch size of 64, hidden, and input size of 50 units. For LSTM and ONLSTM, the initial learning rate is 0.005, while for the GRU and DRNN, it is 0.01. No gradient clipping is performed to train the classifier. 

All models are optimized with Adam \cite{kingma2014adam}, and the codes are written in PyTorch \cite{NEURIPS2019_9015}. 

\subsection{Binary Classifier and Counterfactual Augmentation}
\label{Performance without counterfactual augmentation}

For the binary classifier, we augment each sentence with its corresponding counterfactual example. Augmenting with counterfactual examples is effective in reducing the spurious correlation in sentiment analysis \cite{Kaushik2020Learning}. In our case, the counterfactual example will be constructed by flipping the number of the main verb of a grammatically correct sentence. Thus, we use correct as well as the incorrect version of the same sentence in training. This results in the training size of $195$k sentences for the binary classifier. Table \ref{appen: comp aug vs wo aug} shows the performance with/without counterfactual augmentation. Note that, the accuracy improved substantially for ONLSTM trained on the selectively sampled dataset.

\begin{table*}
\centering
\resizebox{\textwidth}{!}{%
\begin{tabular}{l|r|r|r|r|r|r|r|r}
\multirow{2}{*}{\begin{tabular}[c]{@{}l@{}}Binary Classifier \\ \textbf{Configuration}\end{tabular}} & \multicolumn{2}{c}{LSTM}                                                                                                    & \multicolumn{2}{|c}{GRU}                                                                                                      & \multicolumn{2}{|c}{ONLSTM}                                                                                                   & \multicolumn{2}{|c}{DRNN}                                                                                                      \\ \hline
                                                                             & \begin{tabular}[c]{@{}l@{}}Without\\Augmentation\end{tabular} & \begin{tabular}[c]{@{}l@{}}With \\Augmentation\end{tabular} & \begin{tabular}[c]{@{}l@{}}Without \\Augmentation\end{tabular} & \begin{tabular}[c]{@{}l@{}}With \\Augmentation\end{tabular} & \begin{tabular}[c]{@{}l@{}}Without \\Augmentation\end{tabular} & \begin{tabular}[c]{@{}l@{}}With \\Augmentation\end{tabular} & \begin{tabular}[c]{@{}l@{}}Without \\Augmentation\end{tabular} & \begin{tabular}[c]{@{}l@{}}With \\Augmentation\end{tabular}  \\ \hline
Natural Sampling                                                                      & 0.96                                                          & 0.97                                                        & 0.94                                                           & 0.96                                                        & 0.95                                                           & 0.96                                                        & 0.95                                                           & 0.96                                                         \\ \hline
Selective Sampling                                                                         & 0.64                                                          & 0.64                                                        & 0.62                                                           & 0.67                                                        & 0.59                                                           & 0.69                                                        & 0.71                                                           & 0.74      \\\hline                                                  
\end{tabular}}
\caption{Performance of Binary classifier without counterfactual augmentation. Counterfactual augmentation effectively doubles the training size.}
\label{appen: comp aug vs wo aug}
\end{table*}

\subsection{Performance on Natural Sentences}
\label{app: natural}
In Table \ref{tab:natural} we give a full version of Table \ref{tab:nat} (\S \ref{Performance on Natural Sentences}) including the standard deviations on 5 different runs. 

\subsection{Representation Similarity Analysis}
\label{RSA}
Representation similarity analysis or RSA \cite{abnar2020transferring, laakso2000content} is a technique to analyze the representation level differences among the models. RSA is a standard tool of multivariate statistical analysis used to quantify the relation between the representations. In RSA, we evaluate the second-order similarity to avoid comparing the learned representations from different spaces. For the comparison, we take 2000 natural sentences selected randomly from the test set. A general procedure to perform RSA in 2 dimensions for any number of models is given below:
\begin{enumerate}
    \item For every model, if the hidden states are a matrix of dimension (2000, N), where N is the hidden dimension, then evaluate the similarity matrix as $XX^T$. We then row-wise standard normalize this similarity matrix to get ($\mathscr{S}$).
    \item For every model (i, j) pair of models, we then evaluate the row-wise inner product between $\mathscr{S}_i$ and $\mathscr{S}_j$ and then take an average of all the values to get $\mathscr{M}_{ij}$. 
    \[\mathscr{M}_{ij} = \frac{\sum\limits_{k=1}^{2000} \mathscr{S}_{ik}^T \mathscr{S}_{jk}}{2000} \]
    \item We then perform Multidimensional scaling of the matrix $\mathscr{M}$ to 2 dimensions. This is available in Scikit Learn \cite{JMLR:v12:pedregosa11a}. 
\end{enumerate}

\subsection{Analysis of Variance}
\label{Analysis of Variance}
In this section, we will analyze the spread of the projected representations across different random seeds. Note that, since the projection is in 2 dimensions, to measure the spread evaluate $l_{2}$ norm of a vector of standard deviations across individual components. Table \ref{appen: variance} shows that BC is more susceptible to a local optimum across random initializations.

\begin{table}[ht]
\centering
\resizebox{\linewidth}{!}{%
\begin{tabular}{l|r|r|r|r|r|r}
\multicolumn{1}{l|}{Architecture} & \multicolumn{3}{c|}{Natural Sampling}                                & \multicolumn{3}{c}{Selective Sampling}                     \\ 
\hline
\multicolumn{1}{l|}{} & LM     & BC & \multicolumn{1}{c|}{$\frac{BC}{LM}$} & LM    & BC     & \multicolumn{1}{c}{$\frac{BC}{LM}$}  \\ 
\multicolumn{1}{l|}{} &      &  & \multicolumn{1}{c|}{} &    &     & \multicolumn{1}{c}{}  \\ 

\hline
DRNN                                                             & 44.85  & 168.81     & 3.76                                                    & 40.80 & 458.80 & 11.24                                            \\ 
GRU                                                              & 220.36 & 626.24     & 2.84                                                    & 83.93 & 624.06 & 7.44                                             \\ 
LSTM                                                             & 66.03  & 941.76     & 14.26                                                   & 57.68 & 729.18 & 12.64                                            \\ 
ONLSTM                                                           & 71.48  & 586.42     & 8.20                                                    & 58.41 & 698.10 & 11.95                                            \\
\hline
\end{tabular}}
\caption{L2 Norm of a vector of component-wise standard deviations for both the LM and Binary Classifier (BC). The third column represents the ratio of the two norms for BC and LM. This shows that across random initializations, BC is more susceptible to a local optimum.}
\label{appen: variance}
\end{table}

\subsection{Fine-Tuning}
\label{app fine-tuning}
\citet{lepori-etal-2020-representations} showed that the syntactic robustness of RNNs could be improved by fine-tuning the trained models on a small amount of syntactically challenging data. We consider a similar exercise for our trained language models (Selective sampling), where we further fine-tuned the model with the challenging artificially generated sentences. To avoid a significant shift in the domain of training sentences for the LM, i.e., from the natural sentences to synthetically generated sentences, we avoid adding sentences with agreement across relative clauses. We fine-tune our LM only on prepositional phrases involving one attractor noun. In Table \ref{app:Fine_tuning}, we present the analysis when we fine-tune our trained LMs for 1 and 5 epochs on different fine-tuning set size. 

We notice that the accuracy of sentences without attractor decreases with fine-tuning for all the models, including the one with tree inductive bias.

\begin{table*}[ht]
\centering
\begin{tabular}{c|c|c|c|c|c|c|c} 
\multicolumn{1}{l|}{Architecture} & \multicolumn{1}{l|}{} & \multicolumn{2}{c|}{$\mathscr{N}$=535} & \multicolumn{2}{c|}{$\mathscr{N}$=1069} & \multicolumn{2}{c}{$\mathscr{N}$=1601}  \\ 
\hline
\multicolumn{1}{l|}{}             & No fine-tune              & Epoch=1 & Epoch=5          & Epoch=1 & Epoch=5           & Epoch=1 & Epoch=5            \\ 
\hline
LSTM                               & \textbf{0.90}                  & 0.86    & 0.64             & 0.83    & 0.72              & 0.78    & 0.75               \\ 
ONLSTM                             & \textbf{0.91}                  & 0.87    & 0.48             & 0.84    & 0.44              & 0.78    & 0.60               \\ 
GRU                                & \textbf{0.87}                  & 0.85    & 0.80             & 0.83    & 0.78              & 0.80    & 0.78               \\ 
DRNN                               & \textbf{0.83}                  & 0.80    & 0.77             & 0.78    & 0.76              & 0.77    & 0.78               \\
\hline
\end{tabular}
\caption{Reduction in out of distribution performance with an increasing number of challenging fine-tuning examples and with epochs of re-training. Bolds mark the best performance across the columns for each model.}
\label{app:Fine_tuning}
\end{table*}

\subsection{Comparison with other works based on TSE}
We now compare the LSTM model trained on selectively sampled data with the existing results from \citet{kuncoro2019scalable} on LSTM network distilled with a RNNG \cite{dyer2016recurrent} as a teacher in the Table \ref{appen: tse comparison}. Note, the models by \citet{kuncoro2019scalable} are trained on the Wikipedia dataset made available by \citet{gulordava2018colorless} not on the one made available by \citet{linzen2016assessing}, which we have used to train our models. However, both are extracted from Wikipedia. Only a small model (Small DSA LSTM) was taken to have a fair comparison in terms of dataset size. Small LSTM mentioned by \citet{kuncoro2019scalable} is trained on 600k sentences, while ours is trained on 98k sentences. 

\begin{table*}[ht]
\centering
\resizebox{1\textwidth}{!}{%
\begin{tabular}{l|c|c|c|c|c}
                            & \begin{tabular}[c]{@{}c@{}}Small LSTM \\ (600k) \end{tabular} & \begin{tabular}[c]{@{}c@{}}Small DSA LSTM  \\ (600k) \end{tabular} & \begin{tabular}[c]{@{}c@{}}Our LSTM (Selective) \\ (98k) \end{tabular} & \begin{tabular}[c]{@{}c@{}}Our LSTM (Natural)\\ (98k)\end{tabular} & Humans  \\ \hline
Simple Agreement            & 0.89                                                 & 0.96                                                     & 0.86                                                     & \textbf{0.99 }                                                    & 0.96    \\ 
Short VP                    &\textbf{ 0.90 }                                                 & 0.88                                                     & 0.71                                                     & 0.85                                                     & 0.94    \\ 
Long VP                     & \textbf{0.78 }                                                & 0.74                                                     & 0.69                                                     & 0.65                                                     & 0.82    \\ 
Across Prepositional Phrase & 0.83                                                 & 0.88                                                     &\textbf{ 0.89 }                                                    & 0.86                                                     & 0.85    \\ 
Across Subject RC           & 0.81                                                 & 0.87                                                     & \textbf{0.89}                                                     & 0.81                                                     & 0.88    \\ 
Across Object RC            & 0.54                                                 & 0.69                                                     & \textbf{0.83  }                                                   & 0.73                                                     & 0.85    \\ 
Within Object RC            & 0.79                                                 & \textbf{0.87}                                                    & 0.63                                                     & 0.78                                                     & 0.78    \\ 
Across Object RC (no that)  & 0.55                                                 & 0.61                                                     & \textbf{0.73}                                                     & 0.62                                                     & 0.82    \\ 
Within Object RC (no that)  & 0.72                                                 & \textbf{0.88 }                                                    & 0.61                                                     & 0.72                                                     & 0.79   \\ \hline
\end{tabular}}
\caption{Comparison of our LSTM models with the existing results on the LSTM network distilled with an RNNG as a teacher. Bolds mark the best model for each test in a row; humans are not taken into consideration while bolding. \citet{kuncoro2019scalable} compares the probability of the grammatically correct sentence with its ungrammatical counterpart, whereas we compare the probability of the grammatically correct verb with its ungrammatical counterpart. \citet{kuncoro2019scalable} train their small LSTM LM on a subset of \citet{gulordava2018colorless} Wikipedia corpus.}
\label{appen: tse comparison}
\end{table*}

\subsection{Fine-Grained analysis of TSE}
\label{app fine}
Table \ref{appen table: fine grain analysis-1}, \ref{appen table: fine grain analysis-2} and \ref{appen table: fine grain analysis-3} presents a fine-grained analysis of TSE (\S \ref{Targeted Syntactic Evaluation}) with demarcations based on number of the main subject and the embedded subject. We report the mean and standard deviation of 5 models with different seeds.

\begin{table*}[ht]
\centering
    \resizebox{1\textwidth}{!}{%
\begin{tabular}{l | c c  c c| c c c c} 
\textbf{Architecture}       & \multicolumn{4}{c|}{Natural Sampling}                         & \multicolumn{4}{c}{Selective Sampling}                           \\ 
\hline
      & 0          & 1          & 2          & 3          & 0          & 1          & 2          & 3           \\ 
\hline
      & \multicolumn{8}{c}{LANGUAGE MODEL}                                                                    \\ 
\hline
LSTM   & \textbf{0.98} ($\pm$0.00) & 0.91 ($\pm$0.01) & 0.84 ($\pm$0.03) & 0.78 ($\pm$0.06) & 0.89 ($\pm$0.01) & \textbf{0.98} ($\pm$0.00) & \textbf{0.98} ($\pm$0.00) & 0.95 ($\pm$0.01)  \\ 
ONLSTM &\textbf{0.98} ($\pm$0.00) & 0.92 ($\pm$0.01) & 0.86 ($\pm$0.01) & 0.82 ($\pm$0.03) & 0.90 ($\pm$0.01) & \textbf{0.98} ($\pm$0.00) & \textbf{0.98} ($\pm$0.00) & 0.95 ($\pm$0.01)  \\ 
GRU    & \textbf{0.97} ($\pm$0.00) & 0.88 ($\pm$0.01) & 0.78 ($\pm$0.02) & 0.73 ($\pm$0.03) & 0.87 ($\pm$0.01) & \textbf{0.98} ($\pm$0.00) & 0.97 ($\pm$0.00) & 0.94 ($\pm$0.01)  \\ 
DRNN   & \textbf{0.96} ($\pm$0.00) & 0.69 ($\pm$0.02) & 0.47 ($\pm$0.03) & 0.36 ($\pm$0.03) & 0.83 ($\pm$0.01) & \textbf{0.97} ($\pm$0.00) & 0.94 ($\pm$0.01) & 0.91 ($\pm$0.01)  \\ 
\hline
      & \multicolumn{8}{c}{BINARY CLASSIFIER}                                                                 \\ 
\hline
LSTM   &\textbf{0.97} ($\pm$0.01) & 0.93 ($\pm$0.02) & 0.87 ($\pm$0.03) & 0.82 ($\pm$0.03) & 0.60 ($\pm$0.06) & \textbf{0.98} ($\pm$0.00) & 0.96 ($\pm$0.00) & 0.97 ($\pm$0.01) \\ 
ONLSTM &\textbf{0.97} ($\pm$0.01) & 0.91 ($\pm$0.05) & 0.84 ($\pm$0.07) & 0.81 ($\pm$0.07) & 0.64 ($\pm$0.08) & \textbf{0.98} ($\pm$0.00) & 0.97 ($\pm$0.00) & \textbf{0.98} ($\pm$0.01)  \\ 
GRU    &\textbf{0.97} ($\pm$0.00) & 0.88 ($\pm$0.01) & 0.76 ($\pm$0.02) & 0.69 ($\pm$0.04) & 0.62 ($\pm$0.05) & 0.95 ($\pm$0.01) & 0.94 ($\pm$0.02) & \textbf{0.96} ($\pm$0.01) \\ 
DRNN   & \textbf{0.97} ($\pm$0.00) & 0.90 ($\pm$0.01) & 0.81 ($\pm$0.02) & 0.77 ($\pm$0.02) & 0.70 ($\pm$0.02) &\textbf{0.97} ($\pm$0.00)& 0.96 ($\pm$0.00) & 0.96 ($\pm$0.01)  \\
\hline
\end{tabular}}
\caption{Performance of LM and classifier with an increasing number of attractors between the main subject and verb. Bolds mark the maximum accuracy in each configuration across the attractor, for each model; the more the better.}
\label{tab:natural}
\end{table*}

\begin{table*}[ht]
\centering
\resizebox{\textwidth}{!}{%
\begin{tabular}{l|l|c|c|c|c|c|c|c|c|c}
 Condition                      & Case                                                   & LSTM Natural                                                                      & LSTM Selective                                                                    & ONLSTM Natural                                                                    & ONLSTM Selective                                                                   & GRU Natural                                                                       & GRU Selective                                                                     & DRNN Natural                                                                      & DRNN Selective                                                                       & \# Sentences                                                                                                        \\ 
\hline Simple Agr                     & \begin{tabular}[c]{@{}l@{}}S\\P\end{tabular}           & \begin{tabular}[c]{@{}r@{}}0.98 ($\pm$0.01)\\1.00 ($\pm$0.00) \end{tabular}                              & \begin{tabular}[c]{@{}r@{}}0.73 ($\pm$0.04)\\0.98 ($\pm$0.02) \end{tabular}                           & \begin{tabular}[c]{@{}r@{}}0.98 ($\pm$0.03)\\0.98 ($\pm$0.03) \end{tabular}                              & \begin{tabular}[c]{@{}r@{}}0.73 ($\pm$0.02)\\0.99 ($\pm$0.01) \end{tabular}                            & \begin{tabular}[c]{@{}r@{}}0.97 ($\pm$0.02)\\0.99 ($\pm$0.02) \end{tabular}                              & \begin{tabular}[c]{@{}r@{}}0.77 ($\pm$0.01)\\0.91 ($\pm$0.07) \end{tabular}                           & \begin{tabular}[c]{@{}r@{}}0.98 ($\pm$0.01) \\0.95 ($\pm$0.04)  \end{tabular}                             & \begin{tabular}[c]{@{}r@{}}0.71 ($\pm$0.04)\\0.87 ($\pm$0.09) \end{tabular}                              & \begin{tabular}[c]{@{}r@{}}156\\156 \end{tabular}                                                                   \\ 
\hline
 Short VP                       & \begin{tabular}[c]{@{}l@{}}S\\P\end{tabular}           & \begin{tabular}[c]{@{}r@{}}0.80 ($\pm$0.05)\\0.90 ($\pm$0.06) \end{tabular}                              & \begin{tabular}[c]{@{}r@{}}0.56 ($\pm$0.15)\\0.86 ($\pm$0.08) \end{tabular}                           & \begin{tabular}[c]{@{}r@{}}0.86 ($\pm$0.04)\\0.89 ($\pm$0.04) \end{tabular}                              & \begin{tabular}[c]{@{}r@{}}0.65 ($\pm$0.02)\\0.81 ($\pm$0.17) \end{tabular}                            & \begin{tabular}[c]{@{}r@{}}0.80 ($\pm$0.05)\\0.83 ($\pm$0.08) \end{tabular}                              & \begin{tabular}[c]{@{}r@{}}0.54 ($\pm$0.03)\\0.83 ($\pm$0.06) \end{tabular}                           & \begin{tabular}[c]{@{}r@{}}0.72 ($\pm$0.03) \\0.67 ($\pm$0.08) \end{tabular}                             & \begin{tabular}[c]{@{}r@{}}0.51 ($\pm$0.13)\\0.80 ($\pm$0.06) \end{tabular}                               & \begin{tabular}[c]{@{}r@{}}1716\\1716 \end{tabular}                                                                 \\ 
 \hline
Long VP                        & \begin{tabular}[c]{@{}l@{}}S\\P\end{tabular}           & \begin{tabular}[c]{@{}r@{}}0.40 ($\pm$0.04)\\0.90 ($\pm$0.07) \end{tabular}                              & \begin{tabular}[c]{@{}r@{}}0.52 ($\pm$0.10)\\0.86 ($\pm$0.09) \end{tabular}                           & \begin{tabular}[c]{@{}r@{}}0.42 ($\pm$0.08)\\0.92 ($\pm$0.04) \end{tabular}                              & \begin{tabular}[c]{@{}r@{}}0.39 ($\pm$0.17)\\0.95 ($\pm$0.05) \end{tabular}                            & \begin{tabular}[c]{@{}r@{}}0.40 ($\pm$0.05)\\0.85 ($\pm$0.08) \end{tabular}                              & \begin{tabular}[c]{@{}r@{}}0.32 ($\pm$0.10)\\0.98 ($\pm$0.03) \end{tabular}                           & \begin{tabular}[c]{@{}r@{}}0.30 ($\pm$0.09) \\ 0.83 ($\pm$0.04) \end{tabular}                             & \begin{tabular}[c]{@{}r@{}}0.34 ($\pm$0.06) \\ 0.95 ($\pm$0.04) \end{tabular}                             & \begin{tabular}[c]{@{}r@{}}260\\260 \end{tabular}                                                                   \\ \hline
\end{tabular}}
\caption{Fine-grained experimental results on constructed sentences. Example sentences for each condition are reported in the supplementary material of \citet{marvin2018targeted}. For models trained on selective sampled data configuration, we observe that the performance of the models is far worse for \emph{singular} main nouns than \emph{plural} nouns. Low average performance on sentences with Long Verb Phrase coordination in comparison to other conditions may be attributed to a long-term non-local agreement that needs to be captured by the models. Aggregated results for these conditions are reported in Table \ref{TSE table}.}
\label{appen table: fine grain analysis-1}
\end{table*}

\begin{table*}[ht]
\centering
\resizebox{\textwidth}{!}{%
\begin{tabular}{l|l|c|c|c|c|c|c|c|c|c}
 Condition                      & Case                                                   & LSTM Nat                                                                      & LSTM Selective                                                                    & ONLSTM Natural                                                                    & ONLSTM Selective                                                                   & GRU Natural                                                                       & GRU Selective                                                                     & DRNN Natural                                                                      & DRNN Selective                                                                       & \# Sentences                                                                                                        \\ \hline

Within Obj RC Animate             & \begin{tabular}[c]{@{}l@{}}PP\\PS\\SP\\SS\end{tabular} & \begin{tabular}[c]{@{}r@{}}0.99 ($\pm$0.02)\\0.93 ($\pm$0.08)\\0.54 ($\pm$0.14)\\0.69 ($\pm$0.08) \end{tabular}    & \begin{tabular}[c]{@{}r@{}}0.60 ($\pm$0.10)\\0.54 ($\pm$0.06)\\0.62 ($\pm$0.09)\\0.77 ($\pm$0.06) \end{tabular} & \begin{tabular}[c]{@{}r@{}}1.00 ($\pm$0.00)\\0.98 ($\pm$0.02)\\0.45 ($\pm$0.24)\\0.67 ($\pm$0.15) \end{tabular}    & \begin{tabular}[c]{@{}r@{}}0.71 ($\pm$0.17)\\0.60 ($\pm$0.13)\\0.42 ($\pm$0.10)\\0.62 ($\pm$0.15) \end{tabular}  & \begin{tabular}[c]{@{}r@{}}1.00 ($\pm$0.01)\\0.96 ($\pm$0.05)\\0.44 ($\pm$0.04)\\0.60 ($\pm$0.06) \end{tabular}    & \begin{tabular}[c]{@{}r@{}}0.54 ($\pm$0.03)\\0.53 ($\pm$0.03)\\0.38 ($\pm$0.09)\\0.55 ($\pm$0.04) \end{tabular} & \begin{tabular}[c]{@{}r@{}}0.95 ($\pm$0.07) \\0.89 ($\pm$0.16) \\0.41 ($\pm$0.07) \\0.53 ($\pm$0.08) \end{tabular} & \begin{tabular}[c]{@{}r@{}}0.62 ($\pm$0.17) \\0.58 ($\pm$0.13)\\0.26 ($\pm$0.12)\\0.39 ($\pm$0.11) \end{tabular}   & \begin{tabular}[c]{@{}r@{}}2496\\2496\\2496\\2496 \end{tabular}                                                     \\ 
\hline
 Within Obj RC Inanimate           & \begin{tabular}[c]{@{}l@{}}PP\\PS\\SP\\SS\end{tabular} & \begin{tabular}[c]{@{}r@{}}0.99 ($\pm$0.01)\\0.93 ($\pm$0.09)\\0.49 ($\pm$0.12)\\0.65 ($\pm$0.11) \end{tabular}    & \begin{tabular}[c]{@{}r@{}}0.59 ($\pm$0.09)\\0.53 ($\pm$0.05)\\0.64 ($\pm$0.10)\\0.78 ($\pm$0.06) \end{tabular} & \begin{tabular}[c]{@{}r@{}}1.00 ($\pm$0.00)\\0.98 ($\pm$0.03)\\0.43 ($\pm$0.18)\\0.60 ($\pm$0.16) \end{tabular}    & \begin{tabular}[c]{@{}r@{}}0.72 ($\pm$0.16)\\0.60 ($\pm$0.11)\\0.38 ($\pm$0.05)\\0.65 ($\pm$0.14) \end{tabular}  & \begin{tabular}[c]{@{}r@{}}1.00 ($\pm$0.00)\\0.95 ($\pm$0.05)\\0.43 ($\pm$0.04)\\0.54 ($\pm$0.08) \end{tabular}    & \begin{tabular}[c]{@{}r@{}}0.55 ($\pm$0.02)\\0.53 ($\pm$0.03)\\0.37 ($\pm$0.08)\\0.57 ($\pm$0.05) \end{tabular} & \begin{tabular}[c]{@{}r@{}}0.97 ($\pm$0.05) \\0.91 ($\pm$0.13) \\0.39 ($\pm$0.04) \\0.51 ($\pm$0.10) \end{tabular}  & \begin{tabular}[c]{@{}r@{}}0.66 ($\pm$0.18)\\0.60 ($\pm$0.14)\\0.20 ($\pm$0.09)\\0.38 ($\pm$0.15) \end{tabular}      & \begin{tabular}[c]{@{}r@{}}1008\\1008\\1008\\1008 \end{tabular}                                                     \\ 
\hline
 Within Obj RC Animate (no that)   & \begin{tabular}[c]{@{}l@{}}PP\\PS\\SP\\SS\end{tabular} & \begin{tabular}[c]{@{}r@{}}0.98 ($\pm$0.03)\\0.99 ($\pm$0.02) \\0.45 ($\pm$0.14) \\0.50 ($\pm$0.15) \end{tabular}   & \begin{tabular}[c]{@{}r@{}}0.59 ($\pm$0.08)\\0.54 ($\pm$0.06)\\0.61 ($\pm$0.08)\\0.68 ($\pm$0.07) \end{tabular} & \begin{tabular}[c]{@{}r@{}}1.00 ($\pm$0.00) \\0.99 ($\pm$0.01) \\0.45 ($\pm$0.15) \\0.46 ($\pm$0.17) \end{tabular} & \begin{tabular}[c]{@{}r@{}}0.73 ($\pm$0.17)\\0.65 ($\pm$0.19)\\0.41 ($\pm$0.13)\\0.49 ($\pm$0.11) \end{tabular}  & \begin{tabular}[c]{@{}r@{}}1.00 ($\pm$0.01) \\0.97 ($\pm$0.04) \\0.41 ($\pm$0.07) \\0.49 ($\pm$0.05) \end{tabular} & \begin{tabular}[c]{@{}r@{}}0.63 ($\pm$0.08)\\0.56 ($\pm$0.05)\\0.30 ($\pm$0.14)\\0.39 ($\pm$0.10) \end{tabular} & \begin{tabular}[c]{@{}r@{}}0.98 ($\pm$0.03) \\0.93 ($\pm$0.08) \\0.23 ($\pm$0.08) \\0.36 ($\pm$0.05) \end{tabular} & \begin{tabular}[c]{@{}r@{}}0.66 ($\pm$0.13)\\0.62 ($\pm$0.11)\\0.22 ($\pm$0.13)\\0.29 ($\pm$0.10) \end{tabular}    & \begin{tabular}[c]{@{}r@{}}2496\\2496\\2496\\2496 \end{tabular}  \\ 
\hline
Within Obj RC Inanimate (no that) & \begin{tabular}[c]{@{}l@{}}PP\\PS\\SP\\SS\end{tabular} & \begin{tabular}[c]{@{}r@{}}0.99 ($\pm$0.02) \\0.98 ($\pm$0.03) \\0.44 ($\pm$0.15) \\0.48 ($\pm$0.16) \end{tabular} & \begin{tabular}[c]{@{}r@{}}0.60 ($\pm$0.09)\\0.54 ($\pm$0.06)\\0.61 ($\pm$0.08)\\0.68 ($\pm$0.06) \end{tabular} & \begin{tabular}[c]{@{}r@{}}1.00 ($\pm$0.00) \\0.99 ($\pm$0.02) \\0.42 ($\pm$0.12) \\0.38 ($\pm$0.18) \end{tabular} & \begin{tabular}[c]{@{}r@{}}0.73 ($\pm$0.18)\\0.63 ($\pm$0.16)\\0.37 ($\pm$0.12)\\0.51 ($\pm$0.12) \end{tabular}  & \begin{tabular}[c]{@{}r@{}}0.99 ($\pm$0.01) \\0.97 ($\pm$0.06) \\0.36 ($\pm$0.06) \\0.44 ($\pm$0.08) \end{tabular} & \begin{tabular}[c]{@{}r@{}}0.64 ($\pm$0.08)\\0.56 ($\pm$0.04)\\0.27 ($\pm$0.13)\\0.40 ($\pm$0.07) \end{tabular} & \begin{tabular}[c]{@{}r@{}}1.00 ($\pm$0.01) \\0.97 ($\pm$0.04) \\0.20 ($\pm$0.07) \\0.32 ($\pm$0.03) \end{tabular} & \begin{tabular}[c]{@{}r@{}}0.72 ($\pm$0.13)\\0.65 ($\pm$0.16)\\0.18 ($\pm$0.11)\\0.26 ($\pm$0.12) \end{tabular}    & \begin{tabular}[c]{@{}r@{}}1008\\1008\\1008\\1008 \end{tabular}  \\ \hline

\end{tabular}}
\caption{Fine-grained experimental results on constructed sentences having \emph{agreement inside an object RC}. Example: \emph{The farmer that the parents love swims}. These sentences have local dependency (parents and love in the example) where interference is caused by the first noun (farmer). In the second column, the former character (P/S) indicates the grammatical number of the first noun, and the next character indicates the grammatical number of the main noun against which the agreement is being tested. Other example sentences for each condition are reported in the supplementary material of \citet{marvin2018targeted}. We observe interference effects from the singular initial noun (SP/SS cases) to be more significant than the plural initial noun (PP/PS cases). The results for animate noun case is consistent with inanimate case (First and second rows are consistent, and third and fourth rows are consistent). We also observe that for models trained on naturalistic data configuration, the performance on `Within Object RC' is better than on `Within Object RC without that,' which corroborates the findings of \citet{marvin2018targeted}. However, no such distinction can be concluded for models trained on selectively sampled data configuration.  Aggregated results for these conditions are reported in Table \ref{TSE table}.}
\label{appen table: fine grain analysis-2}
\end{table*}

\begin{table*}[ht]
\centering
\resizebox{\textwidth}{!}{%
\begin{tabular}{l|l|c|c|c|c|c|c|c|c|c}
 Condition                      & Case                                                   & LSTM Natural                                                                      & LSTM Selective                                                                    & ONLSTM Natural                                                                    & ONLSTM Selective                                                                   & GRU Natural                                                                       & GRU Selective                                                                     & DRNN Natural                                                                      & DRNN Selective                                                                       & \# Sentences                                                                                                        \\ \hline
 
 Across Prep Animate               & \begin{tabular}[c]{@{}l@{}}PP\\PS\\SP\\SS\end{tabular} & \begin{tabular}[c]{@{}r@{}}1.00 ($\pm$0.00)\\0.92 ($\pm$0.02)\\0.60 ($\pm$0.10)\\0.91 ($\pm$0.05) \end{tabular}    & \begin{tabular}[c]{@{}r@{}}0.99 ($\pm$0.02)\\1.00 ($\pm$0.00)\\0.84 ($\pm$0.05)\\0.73 ($\pm$0.07) \end{tabular} & \begin{tabular}[c]{@{}r@{}}0.98 ($\pm$0.02)\\0.89 ($\pm$0.06)\\0.69 ($\pm$0.13)\\0.94 ($\pm$0.04) \end{tabular}    & \begin{tabular}[c]{@{}r@{}}0.99 ($\pm$0.01)\\1.00 ($\pm$0.00)\\0.80 ($\pm$0.02)\\0.72 ($\pm$0.01) \end{tabular}  & \begin{tabular}[c]{@{}r@{}}0.99 ($\pm$0.02)\\0.82 ($\pm$0.07)\\0.53 ($\pm$0.09)\\0.91 ($\pm$0.02) \end{tabular}    & \begin{tabular}[c]{@{}r@{}}0.97 ($\pm$0.02)\\0.96 ($\pm$0.02)\\0.82 ($\pm$0.05)\\0.78 ($\pm$0.05) \end{tabular} & \begin{tabular}[c]{@{}r@{}}0.95 ($\pm$0.03) \\0.52 ($\pm$0.12) \\0.37 ($\pm$0.07) \\0.90 ($\pm$0.03) \end{tabular}  & \begin{tabular}[c]{@{}r@{}}0.97 ($\pm$0.02)\\0.96 ($\pm$0.03) \\0.69 ($\pm$0.04) \\0.69 ($\pm$0.03) \end{tabular}  & \begin{tabular}[c]{@{}r@{}}7488\\7488\\7488\\7488 \end{tabular}                                                     \\ 
\hline
 Across Prep Inanimate             & \begin{tabular}[c]{@{}l@{}}PP\\PS\\SP\\SS\end{tabular} & \begin{tabular}[c]{@{}r@{}}1.00 ($\pm$0.00)\\0.86 ($\pm$0.05)\\0.77 ($\pm$0.07)\\0.86 ($\pm$0.01) \end{tabular}    & \begin{tabular}[c]{@{}r@{}}0.98 ($\pm$0.04)\\0.95 ($\pm$0.05)\\0.88 ($\pm$0.04)\\0.96 ($\pm$0.04) \end{tabular} & \begin{tabular}[c]{@{}r@{}}0.99 ($\pm$0.01)\\0.85 ($\pm$0.03)\\0.80 ($\pm$0.07)\\0.86 ($\pm$0.00) \end{tabular}    & \begin{tabular}[c]{@{}r@{}}0.96 ($\pm$0.02)\\0.93 ($\pm$0.03)\\0.95 ($\pm$0.05)\\0.97 ($\pm$0.03) \end{tabular}  & \begin{tabular}[c]{@{}r@{}}0.96 ($\pm$0.03)\\0.86 ($\pm$0.10)\\0.73 ($\pm$0.09)\\0.90 ($\pm$0.04) \end{tabular}    & \begin{tabular}[c]{@{}r@{}}0.96 ($\pm$0.04)\\0.95 ($\pm$0.02)\\0.93 ($\pm$0.04)\\0.93 ($\pm$0.02) \end{tabular} & \begin{tabular}[c]{@{}r@{}}0.92 ($\pm$0.04) \\0.61 ($\pm$0.13) \\0.34 ($\pm$0.25) \\0.88 ($\pm$0.01) \end{tabular} & \begin{tabular}[c]{@{}r@{}}0.93 ($\pm$0.04) \\0.93 ($\pm$0.05) \\0.89 ($\pm$0.01) \\0.89 ($\pm$0.06) \end{tabular} & \begin{tabular}[c]{@{}r@{}}1008\\1008\\1008\\1008 \end{tabular}  \\ 
\hline
 Across Sub. RC                 & \begin{tabular}[c]{@{}l@{}}PP\\PS\\SP\\SS\end{tabular} & \begin{tabular}[c]{@{}r@{}}1.00 ($\pm$0.00)\\0.87 ($\pm$0.07)\\0.42 ($\pm$0.16)\\0.94 ($\pm$0.03) \end{tabular}    & \begin{tabular}[c]{@{}r@{}}0.98 ($\pm$0.04)\\0.99 ($\pm$0.01)\\0.80 ($\pm$0.11)\\0.79 ($\pm$0.08) \end{tabular} & \begin{tabular}[c]{@{}r@{}}1.00 ($\pm$0.01)\\0.86 ($\pm$0.10)\\0.46 ($\pm$0.11)\\0.93 ($\pm$0.06) \end{tabular}    & \begin{tabular}[c]{@{}r@{}}1.00 ($\pm$0.00)\\0.99 ($\pm$0.01)\\0.74 ($\pm$0.05)\\0.76 ($\pm$0.06) \end{tabular}  & \begin{tabular}[c]{@{}r@{}}0.99 ($\pm$0.01)\\0.77 ($\pm$0.11)\\0.37 ($\pm$0.13)\\0.94 ($\pm$0.01) \end{tabular}    & \begin{tabular}[c]{@{}r@{}}0.95 ($\pm$0.03)\\0.94 ($\pm$0.02)\\0.71 ($\pm$0.14)\\0.84 ($\pm$0.05) \end{tabular} & \begin{tabular}[c]{@{}r@{}}0.97 ($\pm$0.03) \\0.27 ($\pm$0.03) \\0.16 ($\pm$0.13) \\0.92 ($\pm$0.03) \end{tabular} & \begin{tabular}[c]{@{}r@{}}0.98 ($\pm$0.02) \\0.94 ($\pm$0.04) \\0.56 ($\pm$0.19) \\0.71 ($\pm$0.05) \end{tabular} & \begin{tabular}[c]{@{}r@{}}2496\\2496\\2496\\2496 \end{tabular}  \\ 
\hline
 Across Obj RC Animate              & \begin{tabular}[c]{@{}l@{}}PP\\PS\\SP\\SS\end{tabular} & \begin{tabular}[c]{@{}r@{}}0.93 ($\pm$0.04)\\0.80 ($\pm$0.08)\\0.41 ($\pm$0.23)\\0.77 ($\pm$0.13) \end{tabular}    & \begin{tabular}[c]{@{}r@{}}0.88 ($\pm$0.08)\\0.78 ($\pm$0.22)\\0.73 ($\pm$0.05)\\0.88 ($\pm$0.08) \end{tabular} & \begin{tabular}[c]{@{}r@{}}0.87 ($\pm$0.11)\\0.76 ($\pm$0.09)\\0.62 ($\pm$0.19)\\0.87 ($\pm$0.07) \end{tabular}    & \begin{tabular}[c]{@{}r@{}}0.89 ($\pm$0.03)\\0.85 ($\pm$0.05)\\0.76 ($\pm$0.02)\\0.86 ($\pm$0.06) \end{tabular}  & \begin{tabular}[c]{@{}r@{}}0.69 ($\pm$0.25)\\0.75 ($\pm$0.11)\\0.69 ($\pm$0.13)\\0.75 ($\pm$0.07) \end{tabular}    & \begin{tabular}[c]{@{}r@{}}0.85 ($\pm$0.06)\\0.85 ($\pm$0.09)\\0.68 ($\pm$0.12)\\0.79 ($\pm$0.03) \end{tabular} & \begin{tabular}[c]{@{}r@{}}0.79 ($\pm$0.17) \\0.45 ($\pm$0.20) \\0.52 ($\pm$0.17) \\0.76 ($\pm$0.09) \end{tabular} & \begin{tabular}[c]{@{}r@{}}0.89 ($\pm$0.04) \\0.81 ($\pm$0.11)\\0.66 ($\pm$0.08) \\0.76 ($\pm$0.05) \end{tabular}  & \begin{tabular}[c]{@{}r@{}}2496\\2496\\2496\\2496 \end{tabular}  \\ 
\hline
  Across Obj RC Inanimate            & \begin{tabular}[c]{@{}l@{}}PP\\PS\\SP\\SS\end{tabular} & \begin{tabular}[c]{@{}r@{}}0.91 ($\pm$0.08)\\0.75 ($\pm$0.14)\\0.50 ($\pm$0.27) \\ 0.79 ($\pm$0.07) \end{tabular}  & \begin{tabular}[c]{@{}r@{}}0.82 ($\pm$0.13)\\0.74 ($\pm$0.28)\\0.88 ($\pm$0.04)\\0.94 ($\pm$0.03) \end{tabular} & \begin{tabular}[c]{@{}r@{}}0.87 ($\pm$0.10)\\0.77 ($\pm$0.12)\\0.73 ($\pm$0.17)\\0.85 ($\pm$0.01) \end{tabular}    & \begin{tabular}[c]{@{}r@{}}0.86 ($\pm$0.05)\\0.83 ($\pm$0.11)\\0.90 ($\pm$0.08) \\0.90 ($\pm$0.08) \end{tabular} & \begin{tabular}[c]{@{}r@{}}0.66 ($\pm$0.31)\\0.71 ($\pm$0.13)\\0.76 ($\pm$0.13)\\0.83 ($\pm$0.04) \end{tabular}    & \begin{tabular}[c]{@{}r@{}}0.84 ($\pm$0.11)\\0.87 ($\pm$0.09)\\0.81 ($\pm$0.10)\\0.88 ($\pm$0.09) \end{tabular} & \begin{tabular}[c]{@{}r@{}}0.73 ($\pm$0.14) \\0.40 ($\pm$0.19) \\0.65 ($\pm$0.16) \\0.81 ($\pm$0.05) \end{tabular} & \begin{tabular}[c]{@{}r@{}}0.83 ($\pm$0.07) \\0.80 ($\pm$0.04) \\0.87 ($\pm$0.08) \\0.93 ($\pm$0.05) \end{tabular}  & \begin{tabular}[c]{@{}r@{}}1008\\1008\\1008\\1008 \end{tabular}  \\ 
\hline
 Across Obj RC Animate (no that)    & \begin{tabular}[c]{@{}l@{}}PP\\PS\\SP\\SS\end{tabular} & \begin{tabular}[c]{@{}r@{}}0.88 ($\pm$0.11) \\0.61 ($\pm$0.13) \\0.25 ($\pm$0.12) \\0.69 ($\pm$0.15) \end{tabular} & \begin{tabular}[c]{@{}r@{}}0.81 ($\pm$0.25)\\0.61 ($\pm$0.24)\\0.62 ($\pm$0.15)\\0.84 ($\pm$0.06) \end{tabular} & \begin{tabular}[c]{@{}r@{}}0.75 ($\pm$0.17) \\0.44 ($\pm$0.13) \\0.50 ($\pm$0.21) \\0.79 ($\pm$0.05) \end{tabular} & \begin{tabular}[c]{@{}r@{}}0.87 ($\pm$0.06)\\0.72 ($\pm$0.11)\\0.69 ($\pm$0.03)\\0.84 ($\pm$0.08) \end{tabular}  & \begin{tabular}[c]{@{}r@{}}0.68 ($\pm$0.24) \\0.50 ($\pm$0.12) \\0.53 ($\pm$0.22) \\0.69 ($\pm$0.10) \end{tabular}  & \begin{tabular}[c]{@{}r@{}}0.78 ($\pm$0.07)\\0.71 ($\pm$0.14)\\0.47 ($\pm$0.14)\\0.76 ($\pm$0.05) \end{tabular} & \begin{tabular}[c]{@{}r@{}}0.84 ($\pm$0.15) \\0.55 ($\pm$0.20) \\0.45 ($\pm$0.19) \\0.73 ($\pm$0.10) \end{tabular}   & \begin{tabular}[c]{@{}r@{}}0.82 ($\pm$0.09)\\0.74 ($\pm$0.11)\\0.63 ($\pm$0.08)\\0.73 ($\pm$0.07) \end{tabular}    & \begin{tabular}[c]{@{}r@{}}2496\\2496\\2496\\2496 \end{tabular}  \\ 
\hline
 Across Obj RC Inanimate (no that)  & \begin{tabular}[c]{@{}l@{}}PP\\PS\\SP\\SS\end{tabular} & \begin{tabular}[c]{@{}r@{}}0.83 ($\pm$0.14) \\0.59 ($\pm$0.11) \\0.43 ($\pm$0.18) \\0.77 ($\pm$0.09) \end{tabular} & \begin{tabular}[c]{@{}r@{}}0.71 ($\pm$0.22)\\0.63 ($\pm$0.28)\\0.82 ($\pm$0.08)\\0.91 ($\pm$0.06) \end{tabular} & \begin{tabular}[c]{@{}r@{}}0.71 ($\pm$0.13) \\0.45 ($\pm$0.20) \\0.67 ($\pm$0.19) \\0.82 ($\pm$0.02) \end{tabular}  & \begin{tabular}[c]{@{}r@{}}0.86 ($\pm$0.05)\\0.74 ($\pm$0.13)\\0.86 ($\pm$0.04)\\0.91 ($\pm$0.05) \end{tabular}  & \begin{tabular}[c]{@{}r@{}}0.63 ($\pm$0.28) \\0.43 ($\pm$0.16) \\0.62 ($\pm$0.21) \\0.78 ($\pm$0.08) \end{tabular} & \begin{tabular}[c]{@{}r@{}}0.73 ($\pm$0.05)\\0.67 ($\pm$0.13)\\0.63 ($\pm$0.04)\\0.88 ($\pm$0.05) \end{tabular} & \begin{tabular}[c]{@{}r@{}}0.81 ($\pm$0.15) \\0.52 ($\pm$0.22) \\0.58 ($\pm$0.20)\\0.81 ($\pm$0.04) \end{tabular}   & \begin{tabular}[c]{@{}r@{}}0.82 ($\pm$0.06)\\0.77 ($\pm$0.14)\\0.84 ($\pm$0.08)\\0.90 ($\pm$0.05) \end{tabular}    & \begin{tabular}[c]{@{}r@{}}1008\\1008\\1008\\1008 \end{tabular}  \\
\hline
\end{tabular}}
\caption{Fine-grained experimental results on constructed sentences having one intervening noun (agreement attractor - SP/PS or non-agreement attractor - PP/SS) between the main noun, and the associated verb. Example sentences for each condition are reported in the supplementary material of \citet{marvin2018targeted}. The performance on sentences with SP/PS cases is better for models trained on naturalistic data configuration than selectively sampled configuration. The results on condition with animate main nouns are not consistent with the ones having an inanimate main noun. The reason behind this inconsistency is still not clear.   Aggregated results for these conditions are reported in Table \ref{TSE table}. }
\label{appen table: fine grain analysis-3}
\end{table*}
\end{document}